\documentclass[sigconf]{acmart}
\usepackage{booktabs} 

\setcopyright{rightsretained}

\setcopyright{acmcopyright}
\acmISBN{978-1-4503-9822-0/22/12}

\acmConference[ICVGIP'22]{13th Indian Conference on Computer Vision, Graphics and Image Processing}{December 2022}{Gandhinagar, India}

\acmBooktitle{Proceedings of the Thirteenth Indian Conference on Computer Vision, Graphics and Image Processing (ICVGIP'22), December 8--10, 2022, Gandhinagar, India}

\acmYear{2022}
\copyrightyear{2022}
\acmPrice{15.00}

\editor{Soma Biswas}
\editor{Shanmuganathan Raman}
\editor{Amit K Roy-Chowdhury}

\acmDOI{10.1145/3571600.3571613}

\acmArticle{13}

\begin{document}
\title{Towards Robust Handwritten Text Recognition with On-the-fly User Participation}
\author{Ajoy Mondal}
\authornote{Both authors contributed equally to this research.}
\affiliation{%
\institution{CVIT, IIIT Hyderabad}
\country{India}
}
\email{ajoy.mondal@iiit.ac.com}
\author{Rohit saluja}
\authornotemark[1]
\affiliation{%
\institution{CVIT, IIIT Hyderabad}
\country{India}
}
\email{rohit.saluja@research.iiit.ac.in}
\author{C. V. Jawahar}
\affiliation{%
\institution{CVIT Lab, IIIT Hyderabad}
\country{India}
}
\email{jawahar@iiit.ac.in}

\renewcommand{\shortauthors}{Mondal et al.}

\begin{abstract}
Long-term OCR services aim to provide high-quality output to their users at competitive costs. It is essential to upgrade the models because of the complex data loaded by the users. The service providers encourage the users who provide data where the OCR model fails by rewarding them based on data complexity, readability, and available budget. Hitherto, the OCR works include preparing the models on standard datasets without considering the end-users. We propose a strategy of consistently upgrading an existing Handwritten Hindi OCR model three times on the dataset of $15$ users. We fix the budget of $4$ users for each iteration. For the first iteration, the model directly trains on the dataset from the first four users. For the rest iteration, all remaining users write a page each, which service providers later analyze to select the $4$ (new) best users based on the quality of predictions on the human-readable words. Selected users write $23$ more pages for upgrading the model. We upgrade the model with Curriculum Learning (CL) on the data available in the current iteration and compare the subset from previous iterations. The upgraded model is tested on a held-out set of one page each from all $23$ users. We provide insights into our investigations on the effect of CL, user selection, and especially the data from unseen writing styles. Our work can be used for long-term OCR services in crowd-sourcing scenarios for the service providers and end users.
\end{abstract}


\maketitle

\begin{figure}
    \centering
    \includegraphics[trim={2cm 6.5cm 7.5cm 0},width=0.95\linewidth]{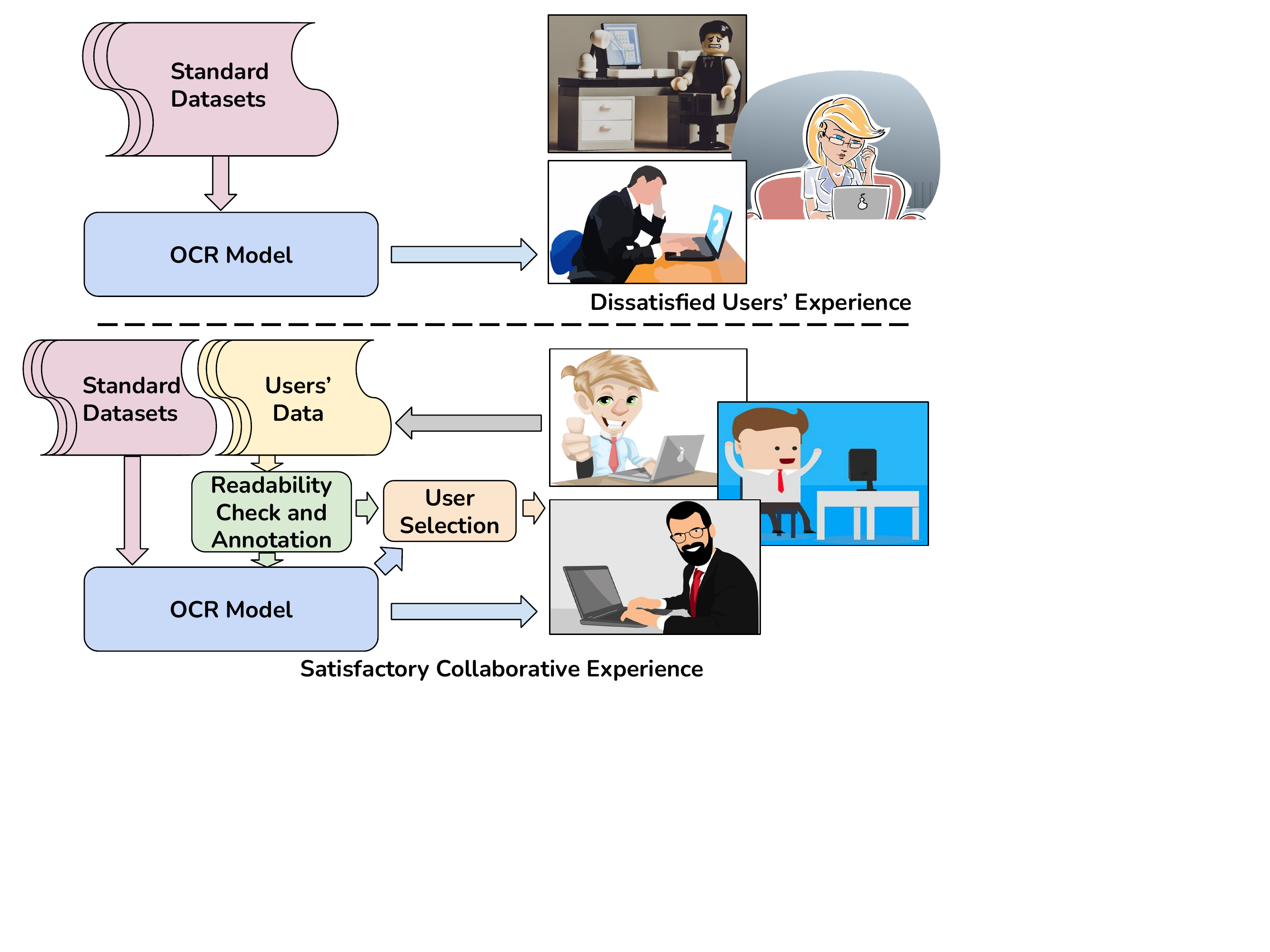}
    \caption{Top: pipeline showing dissatisfied users' experience when the {\sc ocr} service providers and users work in individual silos. Bottom: collaborative pipeline motivating the importance of satisfactory collective experience with on-the-fly user participation in regularly upgrading the {\sc ocr} model.}
    \label{fig:teaser}\vspace{-3mm}
\end{figure}

\section{Introduction}

Optical Character Recognition ({\sc ocr}) is the electronic conversion of printed or handwritten document images into a machine-readable form. {\sc ocr} is an essential component for document image analysis. Typically, an {\sc ocr} system includes two main modules (i) a text detection module and (ii) a text recognition module. A text detection module aims to localize all text blocks within the image, either at the word or line levels. In contrast, the text recognition module strives to understand the text image content and transcribe the visual signals into natural language tokens. The problem of handwritten text recognition is more exciting and challenging than printed text due to uneven variations in handwriting style to the writers, content, and time. A person's handwriting is always unique, and this unique property creates motivation and interest among researchers to work in this imperative and challenging field.

In the field of English handwritten text recognition tasks, several methods are proposed; (i) methods~\cite{chaudhary2021easter,ptucha2019intelligent,yousef2020accurate} based on only convolutional neural networks, (ii) methods~\cite{kang2022pay,diaz2021rethinking,li2021trocr} based on transformer networks, and (iii) methods~\cite{wigington2017data,bluche2017gated,michael2019evaluating,kang2018convolve,bluche2016joint,chowdhury2018efficient,sueiras2018offline} based on {\sc cnn-rnn} architectures. Among almost $7000$ languages around the world, {\sc ocr} systems are mostly available for languages that are of huge importance and strong economic value like English~\cite{graves2008offline,pham2014dropout,li2021trocr}, Chinese~\cite{xie2016fully,wu2017handwritten,peng2022recognition}, Arabic~\cite{maalej2020improving,jemni2022domain}, and Japanese~\cite{ly2018training,nguyen2020semantic}. Among $22$ Indic languages, most of the languages derived from Indic script appear to be at risk of vanishing due to the absence of efforts to preserve them. In many Indic scripts, two or more
characters often combine to form conjuncts, considerably increasing the vocabulary to be tackled by {\sc ocr} systems~\cite{citekey}. These inherent features of Indic scripts make Handwritten Recognizer ({\sc hwr}) more challenging than Latin scripts. In contrast to the 52 unique (upper case and lower case) characters in English, most Indic scripts have over 100 unique basic Unicode characters~\cite{pal2004indian}. Several methods offer solutions to Indic handwritten text recognition tasks. These methods can be categorised into (i) segmentation-free but lexicon-dependent methods~\cite{shaw2008offline,shaw2014combination,kaur2021recognition}, (ii) segmentation dependent methods~\cite{arora2010performance,labani2018novel,alonso2014combining,roy2016hmm}, and (iii) sequence-to-sequence i.e., {\sc cnn-rnn} methods~\cite{adak2016offline,dutta2017towards,dutta2018towards,gongidi2021iiit}.

All the {\sc ocr} systems mentioned above for English and Indian languages perform well when the test set's distribution is similar to the data used to train them. When the test set's distribution deviates from the training set, the performance of these {\sc ocr} systems drastically reduces. This situation often arises in a real-world scenario and often leads to a dissatisfied user experience, as shown at the top of Fig.~\ref{fig:teaser}. In this work, we propose a controlled framework for long-term {\sc ocr} services providing high-quality output to their users at fixed budgets. Robust {\sc ocr} systems must update the models that fail on the complex documents uploaded by the users. As Fig.~\ref{fig:teaser} (bottom) depicts, the regular upgradation of the model collaboratively on users' data and standard datasets can help achieve a satisfactory experience for users as well as service providers. The process also involves user selection based on the available budget for annotation and rewarding the users for the data. We expressly assume that serving a large number of users, say $N$, is the end goal of the {\sc ocr} system. In our setting, we have $N_1=4$ users available at the first iteration of upgrading the model, $N_2=10$ at the second, and $N_3=15$ at the third iteration. We assume that $N_3$ represents $N$ to investigate the subject matter in a controlled setting. We fix the budget of $m=4$ users for each iteration. At the start of the second (and third) iteration, all the new users write a page, which the experts or service providers analyze to select the $m$ (new) best users based on the quality of {\sc ocr} predictions on the human-readable words. Each of the selected $m$ users (or fixed $m$ users for iteration 1) writes $23$ more pages for training the model. We upgrade the model in each iteration with Curriculum Learning ({\sc cl}) and test it on the held-out set consisting of one page each from all the users ($N_3$). We provide insights analysis on the effect of {\sc cl}, user selection, and performance of our system on unseen writing styles. The key contributions of this work are as follows:
\begin{itemize}
\item Our setup is unique compared to the previous {\sc ocr} works which do not include users' satisfaction or participation. We aim to solve the problem in a real-life scenario, where users continuously grow for an {\sc ocr} service, trying to satisfy them with the limited budget of the service provider.
\item Our experiments show that i) {\sc cl} can be effective in continuously upgrading the {\sc ocr} model on users' data, ii) user selection based on {\sc ocr} predictions on human-readable user's data can improve results over other selections with fixed budget, and iii) Observation on the dataset from users which the model has never seen (or trained upon) can help quantify the robustness of the models.
\end{itemize}

\section{Related Work}
{\textbf{OCR Services:}}
{\it Tesseract}~\cite{tesseract17} is an open-source {\sc ocr} system that works for over $100$ languages. The first three versions of {\it Tesseract {\sc ocr}} work on the principle of classical machine learning techniques. A Gaussian Mixture Model (GMM) based classifier identifies characters from features based on the vectors obtained from text boundaries. The subsequent versions of {\it Tesseract} involve line-level deep models. {\it Google Docs {\sc ocr}} works for over $245$ languages~\cite{gDocs15}. Tafti et al.~\cite{tafti2016ocr} perform qualitative and quantitative analysis on {\it Google Docs {\sc ocr}}, {\it Tesseract}, {\it ABBYY FineReader}, and {\it Transym} using $1227$ images from $15$ categories like printed, handwritten, noisy,  multi-oriented, multi-lingual images, etc. While most services work well on printed text, many fail on handwritten and noisy text images. While some of the commercial service providers might be using the users' data to improve their models, none of the {\sc ocr} services mentioned above provide a transparent up-gradation process based on users' inputs like ours with a focus on improving the overall user experience.

{\textbf{Handwritten Text Recognition:}} In the space of English handwritten text recognition, few works have used entirely convolutional neural networks without using any recurrent architectures~\cite{chaudhary2021easter,ptucha2019intelligent,yousef2020accurate}. Few recent networks~\cite{ingle2019scalable,coquenet2020recurrence} use gating mechanisms in {\sc cnn}s to compensate for the dependency on {\sc lstm} cells, known as gated Convolutional Neural Networks ({\sc gcn}). These types of networks outperform fully convolutional Network yet they lag behind {\sc rnn}/Transformer based {\sc ocr} models~\cite{kang2022pay,diaz2021rethinking,li2021trocr}. Recurrent Neural Networks ({\sc rnn}s) are successfully applied to solve Handwritten Text Recognition ({\sc htr}) tasks. LSTM-based models can handle long-term context in sequences. The most common architectures~\cite{wigington2017data,bluche2017gated} are combination of {\sc cnn} and {\sc rnn}, where {\sc cnn} is used for feature extraction from word or line images and {\sc rnn} is used for modeling sequential context. Several works~\cite{michael2019evaluating,kang2018convolve,bluche2016joint,chowdhury2018efficient,sueiras2018offline} use various attention mechanisms to improve the performance of {\sc cnn +lstm}. Recently, Transformer based text recognizers~\cite{kang2022pay,diaz2021rethinking,li2021trocr} achieved state-of-the-art performance. Some of these works use a {\sc cnn}-based backbone with self-attention as encoders to understand document images~\cite{li2021trocr}. 

Officially, there are $22$ languages in India, many of which are used only for communication purposes. Among these languages, Hindi, Bangla, and Telugu are the top three languages in terms of the percentage of native speakers~\cite{krishnan2019hwnet}. In many Indic scripts, two or more characters often combine to form conjuncts which considerably increasing the vocabulary to be tackled by {\sc ocr} systems~\cite{citekey}. These inherent features of Indic scripts make Handwritten Recognizer ({\sc hwr}) more challenging than Latin scripts. Compared to the 52 unique (upper case and lower case) characters in English, most Indic scripts have over 100 unique basic Unicode characters~\cite{pal2004indian}.

Three popular ways of building handwritten word recognizers for Indic scripts are available in the literature. The first one is segmentation-free but lexicon-dependent methods which train on the representation of the whole words~\cite{shaw2008offline,shaw2014combination,kaur2021recognition}. Shaw {\em et al.}~\cite{shaw2008offline} represent word images using a histogram of chain-code directions in the image-strips scanning from left to right by a sliding window as the feature vector. A continuous density Hidden Markov Model ({\sc hmm}) recognizes handwritten Devanagari words. Shaw {\em et al.}~\cite{shaw2014combination} discuss a novel combination of two different feature vectors for holistic recognition of offline handwritten word images in the same direction. The second category of approaches involves segmentation of the characters within the word images and recognition of isolated characters using an isolated symbol classifier such as Support Vector Machine ({\sc svm})~\cite{arora2010performance}, and Artificial Neural Network ({\sc ann})~\cite{labani2018novel,alonso2014combining}. Roy {\em et al.}~\cite{roy2016hmm} segment Bengali and Devanagari word images into the upper, middle, and lower zones, using morphology and shape matching. The symbols in the upper and lower zone are recognized using an {\sc svm}, while a Hidden Markov Model ({\sc hmm}) is used to recognize the characters in the middle zone. Finally, the results from all three zones are combined. This category of approaches suffers from the drawback of using an error-prone script-dependent character segmentation algorithm. The third category of approaches treats word recognition as a sequence-to-sequence prediction task where both the input and output are treated as vector sequences. The aim is to maximize the probability of predicting the output label sequence given the input feature sequence~\cite{adak2016offline,dutta2017towards,dutta2018towards}. Garain {\em et al.}~\cite{garain2015unconstrained} propose a recognizer using Bidirectional Long Short-term Memory ({\sc blstm}) with Connectionist Temporal Classification ({\sc ctc}) layer to recognize unconstrained Bengali offline handwriting words. Adak {\em et al.}~\cite{adak2016offline} use Convolutional Neural Network ({\sc cnn}) integrated with an {\sc lstm} along with a {\sc ctc} layer to recognize Bengali handwritten word. In the same direction, Dutta {\em et al.} proposed {\sc CNN-RNN} hybrid end-to-end model to recognize Devanagari, Bengali~\cite{dutta2017towards}, and Telugu~\cite{dutta2018towards} handwritten words. In one of the works by Gongidi~\cite{gongidi2021iiit}, the authors use Spatial Transformer Network along with hybrid {\sc cnn-rnn} with {\sc ctc} layer to recognize word images in eight different Indic scripts such as Bengali, Gurumukhi, Gujarati, Odia, Kannada, Malayalam, Tamil, and Urdu. The authors use various data augmentation functions to improve recognition accuracy. This category of methods does not require character-level segmentation and is not bound to recognize a limited set of words. 

\textbf{Curriculum Learning:} The Curriculum Learning ({\sc cl}) methods are commonly used for computer vision tasks like object recognition~\cite{hacohen2019power,mousavi2021deep,wang2019dynamic} and object detection~\cite{singhorder,goyal2022detecting,wang2018weakly}. The works mentioned above use {\sc cl} to handle intra-class scale variations, inter-class confusions, and challenges involved in weakly-supervised or semi-supervised training. Our method utilizes the {\sc cl}-based handwriting recognition {\sc ocr} model to collaboratively learn from the datasets of the service providers and continuously update the dataset from the growing number of users.

\begin{figure}[t]
    \centering
    \includegraphics[trim={0cm 6.5cm 11.5cm 0},width=\linewidth]{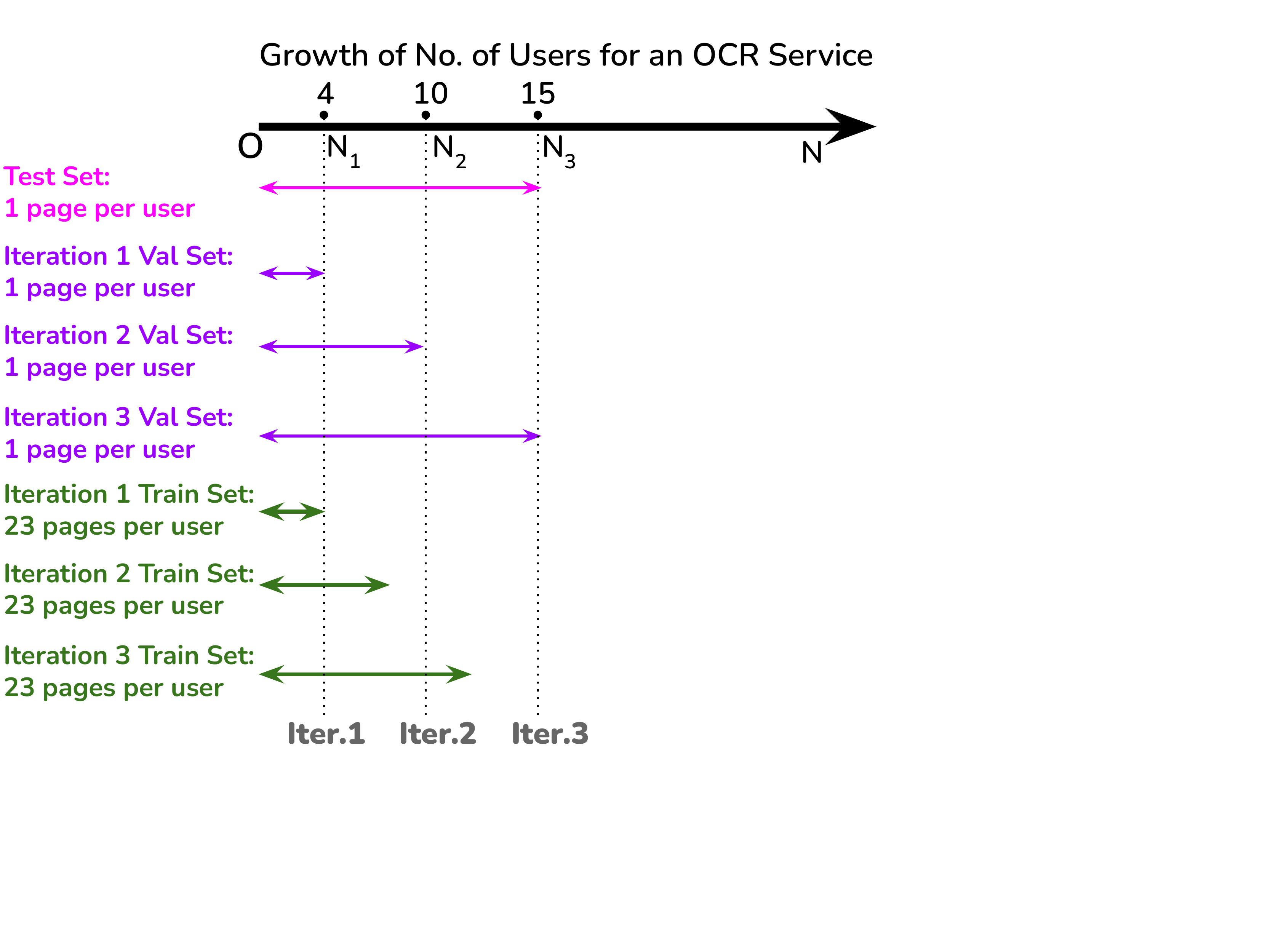}
    \caption{The number of users contributing to the {\sc ocr} service grows with iterations for upgrading the model. Top: test set is fixed initially for $N_3$ users (approximating a large value N). Next: Validation set increases linearly with number of users. Bottom: contribution of users in training set increases relatively lesser compared to validation set with iterations, depending on the budget.}
    \label{fig:data_growth}\vspace{-5mm}
\end{figure}
\begin{figure*}[ht]
    \centering
    \includegraphics[trim={0cm 3cm 0cm 0},width=\linewidth]{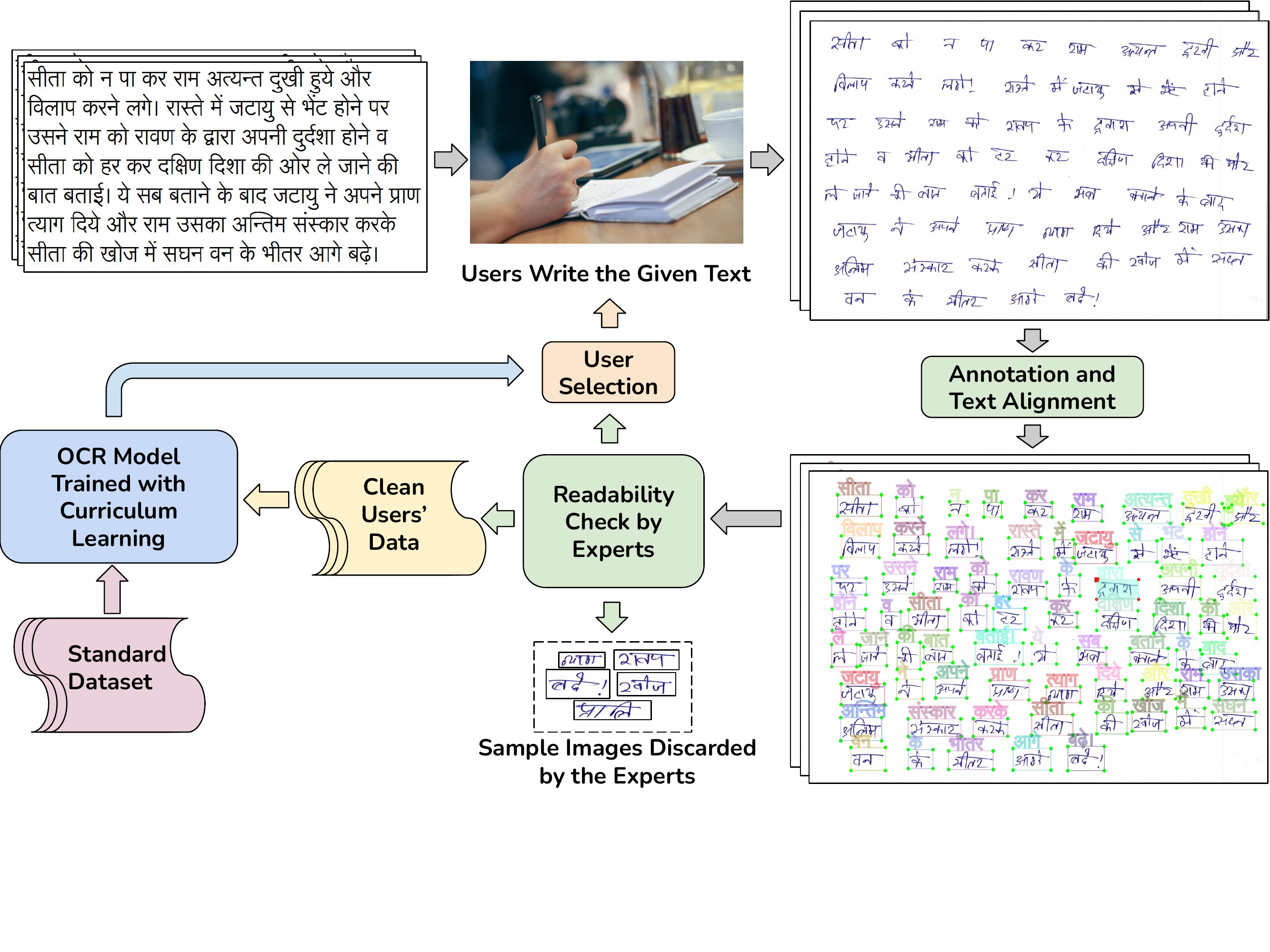}\vspace{-3mm}
    \caption{The pipeline of our framework. Top: Users write the editable text shown on the screen. Bottom-right: annotators label the boxes, and the original text is aligned with the boxes to get word images and labels (shown in different colors). Bottom (middle and left): experts discard the non-readable images, and the {\sc ocr} model is upgraded using the standard and users' datasets.}\vspace{-3mm}
    \label{fig:passive}
\end{figure*}
\section{Methodology}\label{section:system}

This section describes the assumptions and the pipeline of our framework. Consider the problem of updating an {\sc ocr} service on the complex data provided by different users. The number of users generally grows in such scenarios. The end goal of {\sc ocr} service is to provide a satisfactory experience to many users, which we refer to as $N$. However, practically all the users cannot contribute much data but can contribute a few (one or two) pages. Some of the regular users will naturally contribute more data for upgrading the model at arbitrary intervals since such users are also interested in utilizing such a model repeatedly for their documents. The overall situation can be complex in real-life scenarios to the extent that users can arbitrarily grow. The budget of the service providers and the number of pages shared by each can also vary with time. Hence, we make the following assumptions to investigate the work in the controlled setting, which we discuss in the next subsection.
\subsection{Setting}
We define the controlled setting we work with as follows:
\begin{itemize}
    \item As shown in the black arrow on the top of Fig.~\ref{fig:data_growth}, we assume that a limited number of users, i.e., $N_3=15$ approximates the desired value $N$ owing to the limited participants we have. The goal is to develop the robust model for $N_3$ users by using the training data from $<N_3$ users depending on the fixed budget $m=4$ for each of the three training or upgrading iterations.
    \item We fix the test set as one page written from all $N_3$ users as shown in Fig.~\ref{fig:data_growth} (top) in pink.
    \item As Fig.~\ref{fig:data_growth} depicts, we upgrade the models at a fixed set of intervals, i.e., after receiving the data from $N_1=4$, $N_2=10$, and $N_3=15$ users, respectively, which we refer to as three iterations.
    \item As Fig.~\ref{fig:data_growth} (middle) in purple depicts,the validation dataset grows along with the training iteration. The validation set includes one page from each user who is available (partly or completely) until the iteration ends.
    \item The latest model from the previous iteration evaluates the performance of the validation data. Based on the {\sc ocr} performance on each user's validation page, the experts select $m=4$ new users for the current iteration. Here, $m$ is the (fixed) budget signifying the number of users per iteration. We discuss the selection criteria used by experts in the following subsections.
    \item Each of the selected $m$ users write $23$ more pages as shown in Fig.~\ref{fig:data_growth} (middle) in green. The model trained with Curriculum Learning ({\sc cl}) on the dataset from selected users in the current iteration and equal proportions of data from previous iterations. Since the data from users selected in previous iterations are already available, they do not contribute to the training data in the current iteration again. 
\end{itemize}
We discuss the entire pipeline of our system along with the user selection criteria used by the experts in the following subsection.
\subsection{Pipeline}

In Fig.~\ref{fig:passive}, we illustrate the details of training our {\sc ocr} model with Curriculum Learning ({\sc cl}) on a standard dataset and the dataset collected by the users (or writers). We now discuss the collaborative training process in different iterations. 

\begin{figure}[t]
    \centering
    \includegraphics[trim={0 2.5cm 0 0},width=\linewidth]{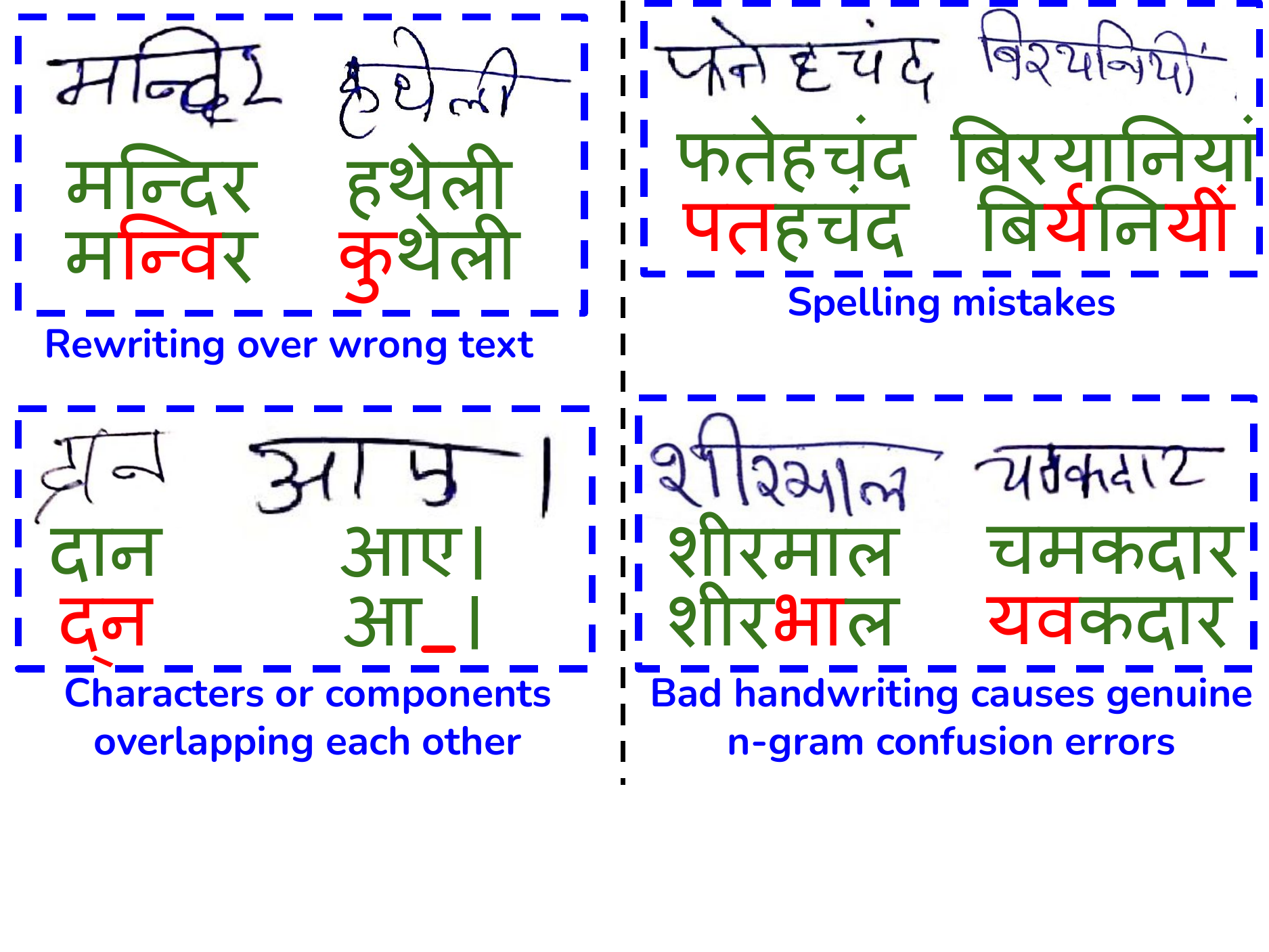}\vspace{-3mm}
    \caption{Hard readable (left) and non-human-readable (right) samples. Each sample is followed by ground truth and predictions (incorrect characters in red, correct in green). The reasons by experts are shown below the dotted boxes.}
    \label{fig:hardvsnonhuman}\vspace{-6mm}
\end{figure}
{\bf Collaborative Training in First Iteration:} In the first iteration of our process, the $N_1$ users write the editable text provided to them, as shown at the top of Fig.~\ref{fig:passive}. As the bottom-right part of the figure depicts, the handwritten documents are then passed to the annotators for adding word-level bounding boxes over the images and aligning the editable text with the bounding boxes.
Finally, experts look at the word images and the corresponding text and discard the non-readable words. The discarded samples are shown in the bottom-middle of Fig.~\ref{fig:passive}. Finally, as the bottom-left of Fig.~\ref{fig:passive} depicts, we update the {\sc ocr} model with {\sc cl} on the equal number of word-level samples from the standard dataset, as obtained from the clean users' data after filtering it from the experts.

\begin{figure}
\centering
\includegraphics[width=\linewidth]{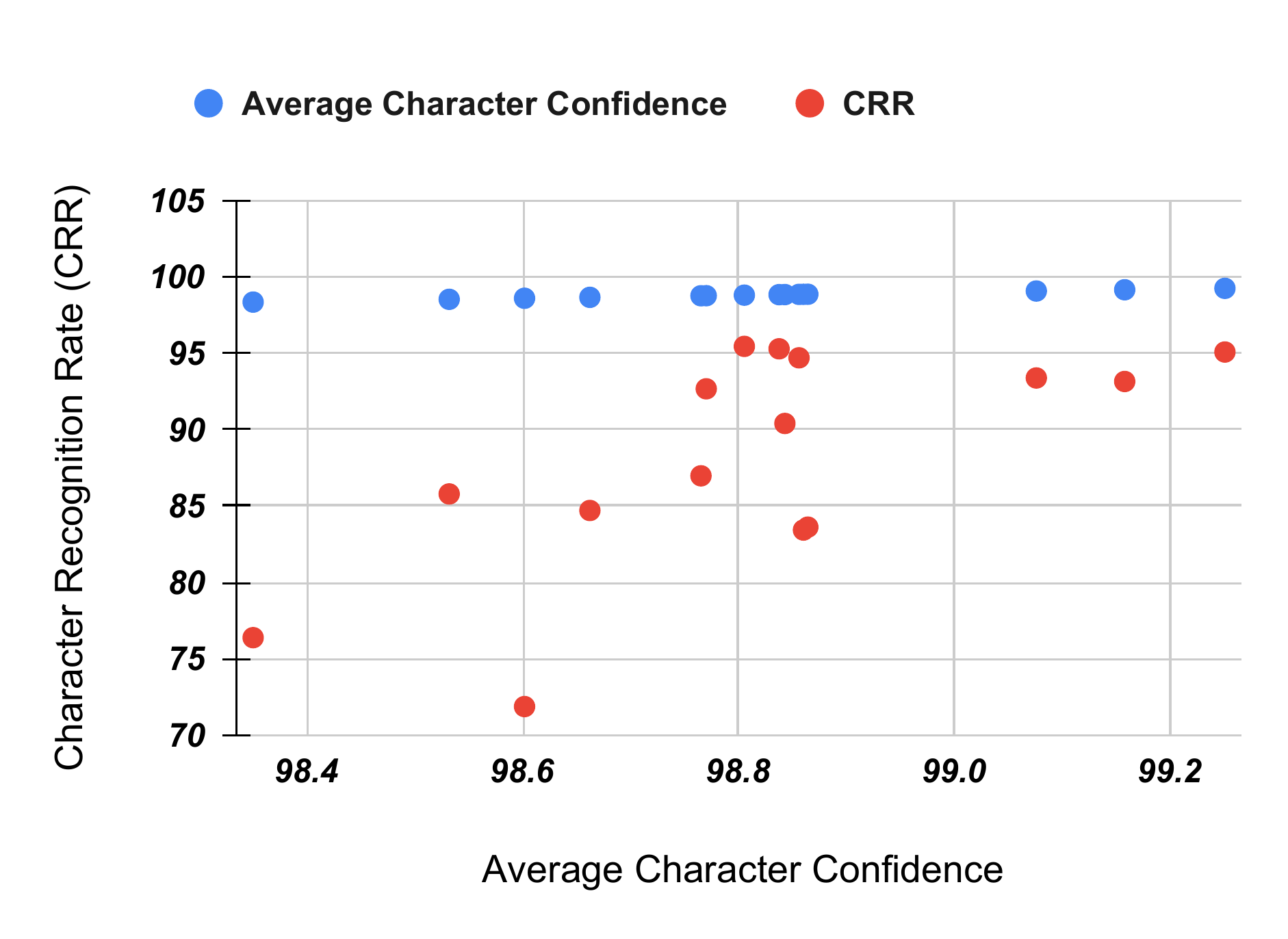}\vspace{-4 mm}
\caption{Scatter Chart: Character Recognition Rate V/S Average Character Confidence of the Iteration 1 model evaluated on the test set from different users. Character Confidence remains high ($> 98.4$) despite varying Recognition Rates.}\vspace{-4 mm}
\label{fig:scatter}
\end{figure}

{\bf Collaborative Training in Subsequent Iterations:}
As discussed in the previous sub-section, for iterations, $t\in[2,3]$, the validation set provided by each new user is analyzed by the experts to finalize the $m$ best users. Particularly, the users with the worst {\sc ocr} accuracy on human-readable samples are selected so that their data can contribute more to the robustness. There is a thin line between the hard readable examples and the non-human-readable examples. Hence, the experts also give the reasons for classifying different samples into readable, hard (but readable), and non-human-readable categories. Hard readable samples from different iterations are shown in Fig.~\ref{fig:hardvsnonhuman} (left), along with the expert reasons (for marking them hard) below the dotted blue boxes. Generally, hard samples contain extreme noise and blur while scanning document images using a phone camera, rewriting certain glyphs over others, and overlapping word or character components. Fig.~\ref{fig:hardvsnonhuman} (right) illustrates the sample non-human-readable samples discarded by the experts. As shown in the text below dotted boxes, non-readable samples involve clear spelling mistakes and bad handwriting cases, which generally lead to Out-Of-Vocabulary (OOV) word predictions. For user selection, experts also analyze the average character confidence over the data from different users. However, as shown in Fig.~\ref{fig:scatter}, we notice that the {\sc OCR} confidence (of the model trained on iteration 1 data) remains high irrespective of the data samples from different users with varying character error rates.
At the end of each iteration $t\in[1,3]$, we update the {\sc ocr} model with {\sc cl} on the $m$ users' dataset obtained from the current iteration and proportionate data\footnote{The number of samples used to update the model from the dataset from iterations $[0, t-1]$ is kept equal to the number of samples obtained in the iteration $t$.} from the previous iterations $[0, t-1]$ (along with the standard dataset denoted by iteration $0$).

\section{Experiments}\label{section:exps}
We use the network architecture proposed by Gongidi et al.~\cite{gongidi2021iiit} for this experiment and refer to its model and dataset as belonging to iteration $0$. The used network consists of four modules: Transformation Network ({\sc tn}), Feature Extractor ({\sc fe}), Sequence Modeling ({\sc sm}), and finally Predictive Modeling ({\sc pm}). The transformation Network has six plain convolutional layers with $16$, $32$, $64$, $128$, $128$, and $128$ channels. Each layer has the filter size, stride, and padding size of $3$, $1$, and $1$, followed by a $2\times2$ max-pooling layer with a stride of $2$. The Feature Extractor module consists of ResNet architecture. The Sequence Modeling consists of a $2$ layer Bidirectional {\sc lstm} ({\sc blstm}) architecture with $256$ hidden neurons in each layer. The predictive Modeling consists of Connectionist Temporal Classification ({\sc ctc}) to decode and recognize the characters by aligning the feature sequence and the target character sequence. We resize input images into $96\times256$. We use Adadelta optimizer for Stochastic Gradient
Descent ({\sc sgd}) for all the experiments. We set the learning rate to $0.001$, batch size to $64$, and momentum to $0.09$. For every curriculum fine-tuning, we reset the learning rate to 0.001. The code and trained model are available at~\footnote{\url{https://github.com/ajoymondal/Towards-Robust-Handwritten-Text-Recognition-with-On-the-fly-User-Participation}}. 

We carry out the following experiments to study the effect of curriculum learning, and user selection:
\begin{itemize}
    \item {\bf Iter1CL:} We compare the Iteration $1$ model trained with i) Curriculum Learning ({\sc cl}) on dataset from iterations $[0,1]$ and ii) without {\sc cl}, i.e., with just Fine Tuning Iteration $0$ model on Iteration $1$ training sets. The former is referred to as {\it Iter1CL} and latter as {\it Iter1FT}.
    \item {\bf Iter2CLm:} For Iteration $2$, we compare the model trained on i) training set from $m$ selected users ({\it Iter2CLm}) and ii) swapping the training set from one and two of the users in the selected set of $m$ users, with a training set of remaining  (non-selected) users available in Iteration $2$. We refer to them as {\it Iter2CLmS1} and {\it Iter2CLmS2}. For {\it Iter2CLmS1} the dataset from a user with the highest {\sc ocr} quality\footnote{We refer to the user whose validation set has the highest {\sc ocr} quality among other users as the best performing user.} among $m$ users is swapped with data from user with the overall highest {\sc ocr} quality on validations set. For {\it Iter2CLmS2}, the dataset from two users with the highest {\sc ocr} quality, among $m$ users are swapped with data from the two users with the overall highest {\sc ocr} quality on validations set.
    \item {\bf Iter3CLm:} Finally, we compare the model upgraded on the data collected in the third iteration {\it Iter3CLm}, with {\it Iter1CL} and {\it Iter2CLm} on overall test set, common seen~\footnote{Seen/unseen throughout the paper means the model has seen/unseen the writer's (or user's) distribution during training. A common seen/unseen test set is the test set from users whose data is seen/unseen in all the iterations.} test set, and common unseen test set, to demonstrate the generalization capability of the final model.
\end{itemize}

\begin{table}
\caption{Results showing effect of the Curriculum Learning on the test set of pages collected from 15 users.} \begin{tabular}{l |l | c c}
\textbf{S.No.} &\textbf{Model} &\textbf{CRR} &\textbf{WRR} \\ \hline
1 & {\it Iter0~\cite{gongidi2021iiit}} & 69.35 & 31.61 \\
2 & {\it Iter1FT} & 88.34  & 70.22 \\
3 & {\it Iter1CL} & {\bf 88.43} & {\bf 70.58} \\ \hline
\end{tabular}
\label{table:results_CL}
\end{table}
\begin{figure*}[ht]
\centering
\includegraphics[trim={3.5cm 14cm 5cm 3.5cm},width=0.55\linewidth]{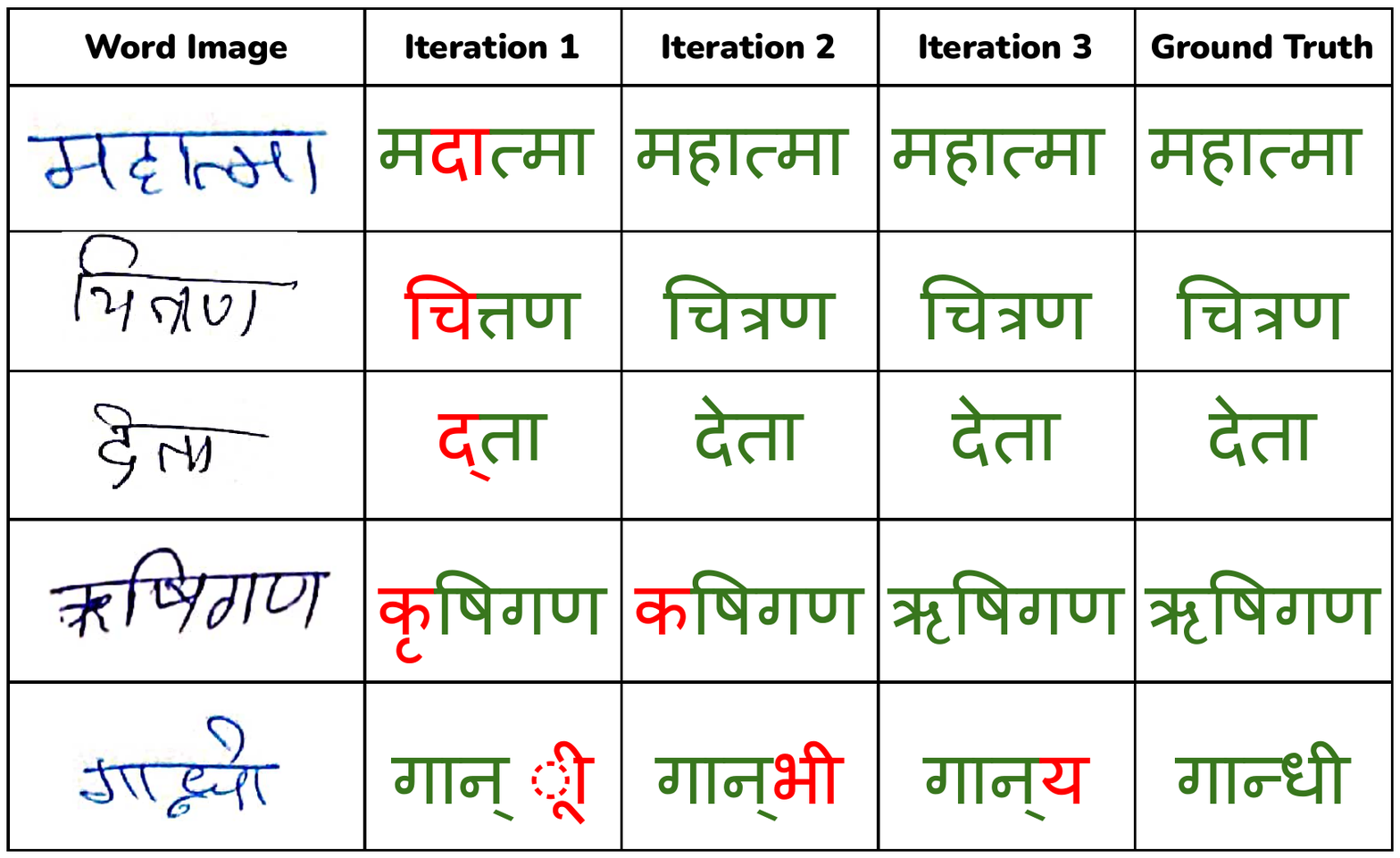}
\caption{Qualitative Results on Unseen Writing Styles of Users with IDs 8, 9, and 15. Rows 1-3: Most of the words are predicted correctly in the second iteration. Row 4: Some rare words are correctly predicted in the third iteration. Row5: A failure case.}
\label{fig:qualitative_Results}
\end{figure*}

\section{Results}\label{section:results}

{\bf Effect of Curriculum Learning:} As shown in the first two rows of Table~\ref{table:results_CL}, just fine-tuning the model by Gongidi et al.~\cite{gongidi2021iiit} on Iteration $1$ dataset, improve the Character Recognition Rate (CRR) and Word Recognition (WRR) significantly by $19\%$ and $38\%$ on the test set. However, Curriculum Learning ({\sc cl}) on the samples collected from the first four users and the equal number of samples from the large training set of Gongidi et al.~\cite{gongidi2021iiit} (words) help us with further improvements, as shown in the third row of Table~\ref{table:results_CL}. As we will see in the next paragraphs, the {\sc cl} also helps us improve results in subsequent iterations.

\begin{table} [ht]
\caption{User selection for iteration 2 based on Human-Readable (HR) words from val. set. Selected users in bold.}
\begin{tabular}{l | c | c | c c}
\textbf{S.No.} &\textbf{User ID} &\textbf{Model} &\textbf{WRR}  &\textbf{WRR-HR} \\ \hline
1 & {\bf 5} &  & 63.11 & 73.03 \\
2 & {\bf 6} &  & 51.06 & 66.67 \\
3 & {\bf 7} & {\it Iter1CL} & 65.12 & 81.25\\
4 & 8 & & 84.61 & 89.19 \\
5 & 9 &  & 80.00 & 84.44  \\
6 & {\bf 10} &  & 43.39 & 55.00 \\ \hline
\end{tabular}
\label{table:UserSel}
\end{table}
{\bf Effect of User Selection:} The results of {\it Iter1CL} model of validation set of Iteration $2$
 from users with IDs $[5,10]$ are shown in Table~\ref{table:UserSel}. Firstly, as shown in the last two columns of the table, the WRR on Human-Readable (WRR-HR) words is generally higher than the WRR on all the words (WRR) in the validation set. However, the sorting order of the recognition rates may change if we consider all words against only  Human-Readable (HR) words. Selecting users based on the worst WRR can have an adverse effect on training the model, since we cannot expect the model to perform well on non-readable samples with spelling mistakes and genuine n-gram confusion errors that occur due to awful handwriting (refer Fig.~\ref{fig:hardvsnonhuman}). Based on WRR-HRR, users with IDs $\{5, 6, 7, 10\}$ are selected for training the {\it Iter2CL} model.
 
\begin{table} [ht]
\caption{Results showing effect of User Selection on test set of pages collected from 15 users.}
\begin{tabular}{l |l | c c}
\textbf{S.No.} &\textbf{Model} &\textbf{CRR} &\textbf{WRR} \\ \hline
1 & {\it Iter2CLmS2} &89.34 &73.12 \\
2 & {\it Iter2CLmS1} &90.49  &75.85 \\
3 & {\it Iter2CLm} &{\bf 91.25} &{\bf 78.43} \\ \hline
\end{tabular}
\label{table:results_UserSel}
\end{table}
The results of model trained on the selected users (with IDs $\{5, 6, 7, 10\}$) are shown in the third row of Table~\ref{table:results_UserSel}. As shown, the user selection based on WRR-HR performs best as compared to the other two variants (rows $1-2$ of Table~\ref{table:results_UserSel}). {\it Iter2CLmS1} involves swapping the best performing selected user's (i.e., user $7$) training set with the overall best performing user's (i.e., user $8$) training set. {\it Iter2CLmS1} achieves slightly degraded performance with a drop of around $1\%$ in CRR and $2\%$ in WRR compared to {\it Iter2CLm} as shown in the last two rows of Table~\ref{table:results_UserSel}. As the first row of the table depicts, the performance drops further in similar proportions if we swap the training data from users $\{5,7\}$, with training data from users $\{8,9\}$.

\begin{table} [ht]
\caption{User selection for iteration 3 based on Human-Readable (HR) words from val. set. Selected users in bold.}
\begin{tabular}{l | c | c | c c}
\textbf{S.No.} &\textbf{User ID} &\textbf{Model} &\textbf{WRR}  &\textbf{WRR-HR} \\ \hline
1 & 8 &  & 84.61 & 89.19 \\ 
2 & 9 &  & 90.00 & 93.62 \\ 
3 & {\bf 11} &  & 77.36 & 79.59\\
4 & {\bf 12} & {\it Iter2CLm} & 68.52 & 82.22 \\
5 & {\bf 13} &  & 84.78 & 88.64 \\ 
6 & {\bf 14} &  & 83.72 & 85.36 \\ 
7 & 15 &  & 82.50 & 93.94  \\ \hline
\end{tabular}
\label{table:iter3sel}
\end{table}

{\bf Results on different Iterations:}
Since the users with IDs $\{8,9\}$ were not selected in the second iteration, we reconsider them in the third iteration. Interestingly, as shown in Table~\ref{table:iter3sel}, the performance of {\it Iter2CLm} model on (unseen) user $8$ validation set is retained compared to {\it Iter1CL} model (Table~\ref{table:results_UserSel} row 4, and the performance on (unseen) user $9$ data improve significantly by $9\%$ (compare row 5 of Table~\ref{table:results_UserSel} with row 2 of Table~\ref{table:iter3sel}). Another interesting highlight of Table~\ref{table:iter3sel}) is that WRR-HR of {\it Iter2CLm} model on all the unseen validation sets from users with IDs $\{8, 9, 11, 12, 13, 14, 15\}$ is close to or above $80\%$. This shows that {\it Iter2CLm} has generalized well. Based on WRR-HR, we select users with IDs $\{11, 12, 13, 14\}$ for training the model in iteration 3.

\begin{table} [ht]
\caption{Results of model trained at different Iterations on overall test set, common seen test set (from users $\{1,2,3,4\}$), and common unseen test set (from users $\{8,9,15\}$).} \label{tab:results}
\begin{tabular}{l |c c | c c | c c }
\textbf{Model} &\multicolumn{2}{|c}{\textbf{All Test Set}} &\multicolumn{2}{|c}{\textbf{Seen Test Set}} &\multicolumn{2}{|c}{\textbf{Unseen Test Set}} \\ \cline{2-7}
    &\textbf{CRR} &\textbf{WRR} &\textbf{CRR} &\textbf{WRR} &\textbf{CRR} &\textbf{WRR} \\ \hline
{\it Iter1CL} &88.43 &70.58 &94.26 &80.44 &92.53 &79.33  \\
{\it Iter2CLm} &91.25 &78.43 &94.29 &83.62 &95.04 &85.28  \\
{\it Iter3CLm} & {\bf 92.32} & {\bf 79.04} &{\bf 94.89} &{\bf 84.29} &{\bf 95.76} &{\bf 85.68} \\ \hline
\end{tabular}
\label{table:diff_iter}
\end{table}

The first three rows of Table~\ref{table:diff_iter} reestablish that {\it Iter2CLm} has generalized well over the overall test set. A similar trend is observed on the seen test set from the users with ids $\{1,2,3,4\}$ who are common seen users across the three iterations, and the unseen test set from users with ids $\{8,9,15\}$ who are common unseen users across the three iterations. The CRR and WRR improve significantly by around $2-3\%$ and $6-8\%$ compared to the iteration $1$ model for the overall test set and standard unseen test set as shown in rows 1-2, columns 2, 3, 6, and 7 of Table~\ref{table:diff_iter}. However, as shown in rows 1-2 and columns 4-5 of the table, the performance gains are slightly lesser (around $0\%$ in CRR and $3\%$ in WRR) on the commonly seen test set. The slight performance gains of {\it Iter3CLm} over {\it Iter2CLm} by $<1\%$ as shown in the last row of  Table~\ref{table:diff_iter} on all the three types of test sets, show that we have reached close to the saturation performance on the test set of users $1-15$. The first three rows in Fig.~\ref{fig:qualitative_Results} for qualitative results on the unseen dataset from users $\{8,9,15\}$ also show that the performance of  {\it Iter2CLm} and {\it Iter3CLm} is similar. One of the rare words corrected in the third iteration and a failure case is shown in the last two rows of Fig.~\ref{fig:qualitative_Results}. To get a rough idea, one can also compare the performance of {\it Iter3CLm} with the CRR and WRR of $93.98$ and $75.41$ reported by Gongidi et al.~\cite{gongidi2021iiit}, wherein the model has seen the writing style of all the users in the test set. Overall, one may argue that the writing style of common unseen users in our case is similar to the seen users, and that is the reason that the accuracy of the unseen test set is high. However, we reason that the high recognition rates on the unseen test set are due to the user selection process described in the previous sections. The selection process takes care that the users with high validation accuracies are less likely to improve the model's generic performance. Hence such users do not contribute more to the training data in subsequent iterations and are filtered out as common unseen users.

\section{Conclusion}
We proposed a controlled framework for upgrading the handwritten {\sc ocr} model on the dataset provided by $15$ users. We upgraded the model three times with a fixed budget of $23$ pages, each from four different users in each iteration. Our work lays the foundation for long-term {\sc ocr} services in crowd-sourcing user scenarios. We believe that our experiments have shown the effectiveness of Curriculum Learning ({\sc cl}) in regular upgradation of the model and user selection under fixed budgets. We finally provide a way to quantify the robustness of {\sc ocr} models using the data with writing style from unseen users. In the future, we would like to explore our work in a real crowd-sourcing scenario in multiple Indian languages, along with numerous factors and a model for user selection.

\bibliographystyle{ACM-Reference-Format}

\end{document}